\documentclass[sigconf]{acmart}

\AtBeginDocument{%
  \providecommand\BibTeX{{%
    \normalfont B\kern-0.5em{\scshape i\kern-0.25em b}\kern-0.8em\TeX}}}

\usepackage{algorithm}
\usepackage{algorithmic}
\usepackage[font=footnotesize]{subfig}
\begin{document}

\title{An Introduction to Robust Graph Convolutional Networks}

\author{Mehrnaz Najafi}
\email{mnajaf2@uic.edu}

\affiliation{%
  \institution{University of Illinois at Chicago}
  \city{Chicago}
  \state{IL}
}
\author{Philip S. Yu}
\email{psyu@uic.edu}

\affiliation{%
  \institution{University of Illinois at Chicago}
  \city{Chicago}
  \state{IL}
}

\begin{abstract}
  Graph convolutional neural networks (GCNs) generalize tradition convolutional neural networks (CNNs) from low-dimensional regular graphs (e.g., image) to high dimensional irregular graphs (e.g., text documents on word embeddings). Due to inevitable faulty data collection instruments, deceptive data manipulation, or other system errors, the data might be error-contaminated. Even a small amount of error such as noise can compromise the ability of GCNs and render them inadmissible to a large extent. The key challenge is how to effectively and efficiently employ GCNs in the presence of erroneous data. In this paper, we propose a novel Robust Graph Convolutional Neural Networks for possible erroneous single-view or multi-view data where data may come from multiple sources. By incorporating an extra layers via Autoencoders into traditional graph convolutional networks, we characterize and handle typical error models explicitly. Experimental results on various real-world datasets demonstrate the superiority of the proposed model over the baseline methods and its robustness against different types of error.
\end{abstract}



\keywords{Natural Language Processing, Deep Learning, Robustness, Multi-View Data, Single-View Data}


\maketitle

\section{Introduction}
Traditional Convolutional Neural Networks (CNN) \cite{lecun-gradientbased-learning-applied-1998} have shown promising capability to extract meaningful features in big datasets, where observations are represented by regular graphs (or grids) such as image \cite{8271995}, video \cite{Kang_2016_CVPR} and speech \cite{conf/interspeech/Abdel-HamidDY13}. CNNs obtain the features by identifying the local shared properties across the data via localized convolutional filters or kernels. Although the preliminary CNN's kernels are hand-engineered, recent CNNs learn them from the data. 

 Text documents on word embeddings, social networks or biological networks that can be represented as graphs lying on irregular or non-Euclidean domains has emerged as a topic of critical significance among the data mining and machine learning community. However, using CNNs for the graphs with irregular structure is not trivial as the convolution and pooling operators are only defined for regular graphs such as grids. To generalize CNNs for irregular graphs effectively and efficiently, Defferrard et al. \cite{Defferrard:2016:CNN:3157382.3157527} proposed a formulation of CNNs, named as Graph Convolutional Neural Networks (GCNs) in the domain of spectral graph theory, which is suitable for both regular and irregular graphs \cite{Chung:1997}. Precisely, GCNs are based on spectral graph theoretical formulation of CNNs on graphs built on graph signal processing with filters with low computational complexity \cite{6494675}.
 
\begin{figure}
  \includegraphics[width=0.3\textwidth]{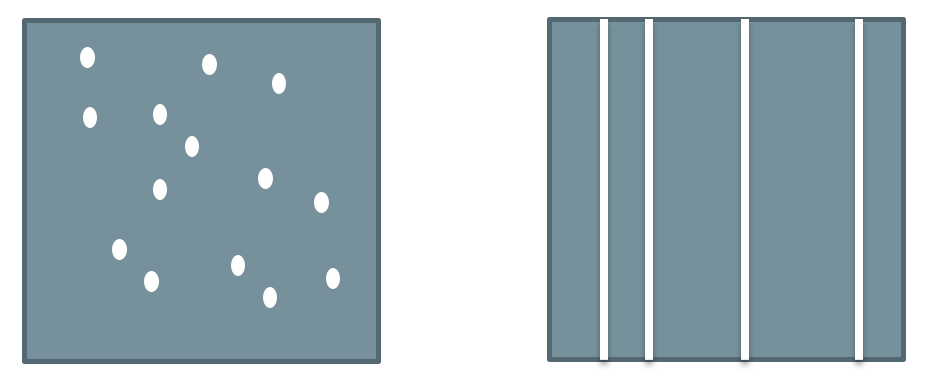}
  \caption{Two types of error (what we show is a erroneous data matrix whose rows are instances and columns are features). Left: Noise Right: Feature Specific Corruptions}
  \Description{Enjoying the baseball game from the third-base
  seats. Ichiro Suzuki preparing to bat.}
  \label{fig:noise-fsc}
\end{figure}
 
 A challenging problem may arise when the data is contaminated by error due to sensor failures, malicious tampering, or API limitation. Error could exist in either the data or label if any available. It could exhibit as \textit{noise} which is a slight perturbation of random subset of entries in data, or \textit{feature-specific corruptions} corresponding to perturbation of a set of random features \cite{Liu:2013:RRS:2412386.2412936}. Fig. \ref{fig:noise-fsc} illustrates these two types of error. With noise and feature-specific corruptions, although error exists only on portions of data, it can adversarially affect the capability of GCNs to learn from data. Consequently, they suffer from performance degradation when applied on erroneous data. For this reason, error-robust GCNs are highly desired. 
 
 Existing work on deep error-robust learning can be roughly classified into four categories: \textit{incorporating extra layers}, \textit{modifying the loss function}, \textit{cleaning up data}, and \textit{enhancing deep learning model capability by using adversarial training}. Adding extra layers to deep learning methods accounts for either error model e.g., noise transition layer or cleaning up data \cite{Han2018,haowang2019}, while modifying the loss function is done either by replacing it with error-tolerant loss or changing the loss with regularization biases \cite{li2016,Zhou:2017:ADR:3097983.3098052,Pu2019LearningLR}. 
 
 The basic idea of cleaning up the data is to first filter out those data instances that tend to be error-contaminated with high confidence and then trains the deep learning model with only the remaining ones \cite{JiangZLLF18}. This approach is often done in a two-step manner (sequentially). Adversarial examples leads deep learning models to learn superficial data statistics and causes significant risk for the models deployed in safety-critical systems such as user authentication. They are often created by injecting noise into clean data instances. Adversarial training improves robustness of the deep learning models against adversarial examples by training the models with them \cite{li2016}.
 
 To improve robustness of GCN against error in data, we propose a novel Robust Graph Convolutional Neural Networks (RGCN) based on GCN. The key idea is that unlike GCN, the proposed RGCN model uses extra layer in the form of robust non-linear deep autoencoders to deal with erroneous data elegantly. We employ two variants of autoencoders: \textit{denoising autoencoders} and \textit{robust low-rank autoencoders}. With the assumption of clean training data, the denoising autoencoders inject error into the data and learn from the erroneous data, while robust low-rank autoencoders assume possibly erroneous data and remove sparse error from it and obtain its clean low-rank approximation. 
 
 The decomposition of possibly erroneous data into clean data and sparse error in robust low-rank autoencoders facilitates robust recovery of low-rank clean component as well as extraction of sparse error component. To capture error as sparse part, the proposed RGCN model imposes $\ell_{1}$ norm on error component. With the popularity of multi-view data, where data may come from multiple sources, we also propose a novel generalization of RGCN for multi-view data (MVRGCN). For example, in Wikipedia, concept of cat may be represented by various views in the form of image (view 1), text (view 2), or even audio (view 3) and videos (view 4). 
 Our main contributions can be summarized as follows:
 
 \begin{itemize}
     \item To the best of our knowledge, the proposed RGCN and MVRGCN models are the first work that recovers clean data via robust low-rank autoencoders, while capturing error for erroneous data. 
     
     \item The proposed RGCN and MVRGCN models is the first method based on low-rank decomposition that isolates the clean low-rank approximation from error for erroneous data.
     
     \item Through extensive experiments on real-world datasets, we show that the proposed RGCN and MVRGCN are superior to several state-of-the-art methods in supervised and semi-supervised learning and robust against error in text and image datasets.
 \end{itemize}

\section{Robust Graph Convolutional Neural Network}
We begin with some necessary notations and concepts of Graph Convolutional Neural Networks (GCNs) and Robust Autoencoders. Table \ref{tab:notation} lists basic symbols that will be used throughout the paper.

\begin{table}
\begin{tabular}{lp{0.35\textwidth}}
\toprule
Symbol & Definition and description\\
\midrule
$x$ & lowercase letter represents a scale\\
$\mathbf{x}$ & boldface lowercase letter represents a vector\\
$\mathbf{X}$ & boldface uppercase letter represents a matrix\\
$\left\|\mathbf{X}\right\|_{F}$ & (Frobenius) norm of matrix $\mathbf{X}$\\
$\left\|\mathbf{X}\right\|_{2}$ & $\ell_2$-norm of matrix $\mathbf{X}$\\
$\left\|\mathbf{X}\right\|_*$ & sum of the singular values of $\mathbf{X}$\\
\bottomrule
\end{tabular}
\caption{List of basic symbols}
\label{tab:notation}
\end{table}


\subsection{Graph Convolutional Neural Network}
Graph Convolutional Neural Networks (GCNs) \cite{Defferrard:2016:CNN:3157382.3157527} is a generalization of traditional Convolutional Neural Networks (CNNs) that can be efficiently and effectively used for graphs with irregular structure. Examples include documents on word embeddings, social networks, and gene data networks. Different from traditional CNNs, GCNs employ localized graph convolution in the spectral domain via Laplacian matrix and graph Fourier transform (GFT). Normalized $\mathbf{L}$ is defined as $\mathbf{L} = \mathbf{I}_n  - \mathbf{D}^{-\frac{1}{2}} \mathbf{W} \mathbf{D}^{-\frac{1}{2}}$, where $\mathbf{W} \in \mathbb{R}^{n \times n}$ ($n$ refers to number vertices in the graph), $\mathbf{D} \in \mathbb{R}^{n \times n}$ is the diagonal matrix with $d_{i, i} = \sum_j w_{i, j}$, and $\mathbf{I} \in \mathbb{R}^{n \times n}$ is the identity matrix.

Since $\mathbf{L}$ is a real symmetric positive semidefinite matrix, it can be decomposed into $\mathbf{L} = \mathbf{U} \mathbf{\Lambda} \mathbf{U}^\top$, where $\mathbf{U} \in \mathbb{R}^{n \times n}$ is the matrix of eigenvectors with $\mathbf{U} \mathbf{U}^\top = \mathbf{I}$, and $\mathbf{\Lambda} \in \mathbb{R}^{n \times n}$ is the diagonal matrix of eigenvalues. Let $x \in \mathbb{R}^n$ be a signal defined on the vertices of the graph, where $x_i$ indicates the value of the signal at the $i$-th vertex. The GFT is obtained via $\hat{\mathbf{x}} = \mathbf{U}^\top \mathbf{x}$, and it converts signal $\mathbf{x}$ to the spectral domain spanned via the Fourier basis $\mathbf{U}$. Based on these concepts, graph convolution can be formulated as follows \cite{Defferrard:2016:CNN:3157382.3157527}:

\begin{align}
\mathbf{y} = g_{\theta}(\mathbf{L})\mathbf{x} = g_{\theta}(\mathbf{U} \mathbf{\Lambda} \mathbf{U}^\top)\mathbf{x} = \mathbf{U} g_{\theta}(\mathbf{\Lambda})\mathbf{U}^\top \mathbf{x}
\label{convolution}
\end{align}

\noindent where the parameter $\theta$ is a vector of Fourier coefficients, and $g_\theta$ is the filter which is a function of $\mathbf{\Lambda}$. For fast filtering, $g_\theta$ can be approximated by Chebyshev polynomials of order $s$ as follows \cite{Defferrard:2016:CNN:3157382.3157527}:

\begin{align}
    g_\theta(\mathbf{\Lambda}) = \sum_{p = 0}^{s - 1} \theta_p T_p(\hat{\mathbf{\Lambda}})
    \label{chebyshev}
\end{align}

\noindent where the parameter $\theta_p \in \mathbb{R}^{s}$ indicates a vector of Chebyshev coefficients and $T_p(\hat{\mathbf{\Lambda}})$ refers to the Chebyshev polynomial of order $s$ evaluated at $\hat{\mathbf{\Lambda}} = \frac{2\mathbf{\Lambda}}{\lambda_{max} - \mathbf{I}}$. By substituting Eq. (\ref{chebyshev}) into Eq. (\ref{convolution}) results in $\mathbf{y} = g_\theta(\mathbf{L})\mathbf{x} = \sum_{p = 0}^{s - 1} \theta_p T_p(\hat{\mathbf{L}}) \mathbf{x}$, where $\hat{\mathbf{L}} = \frac{2}{\lambda_{max}}\mathbf{L} - \mathbf{I}$. We can use the recurrence relation to compute $\hat{x_i} = 2 \hat{\mathbf{L}}\hat{x}_{p-1} - \hat{x}_{p-2}$, if we denote $\hat{x}_p = T_p(\hat{\mathbf{L}})\mathbf{x}$, $\hat{\mathbf{x_0}} = \mathbf{x}$ and $\hat{\mathbf{x}_1} = \hat{\mathbf{L}}\mathbf{x}$. For learning filter, the $j^{th}$ output feature map of the sample $s$ can be obtained as follows \cite{Defferrard:2016:CNN:3157382.3157527}:

\begin{align}
    y_{s, j} = \sum_{i = 1}^{F_{in}} g_{\theta_{i, j}}(\mathbf{L})x_{s, i} \in \mathbb{R}^{n}
\end{align}

\noindent where $x_{s, i}$ denotes the input feature maps and $F_{in} \times F_{out}$ are vectors of Chebyshev coefficients $\theta_{i, j} \in \mathbb{R}^{K}$ refer to the layer's trainable parameters.
Like pooling in traditional CNNs, pooling for GCNs requires neighborhoods on graphs, where similar vertices are grouped together. Likewise, multi-scale clustering of the graph can be used for pooling for multiple layers in GCNs. 
The two final layers in GCN are fully connected layer with an $\ell_2$ regularization on the weights and softmax regression. The loss function for GCN include cross entropy with an $\ell_2$ regularization.


\subsection{Robust Autoencoders}
An autoencoder is a neural network which is trained to attempt to copy its input to its output. Precisely, it learns a mapping from the input to itself through a pair of encoding and decoding phases as follows:
 
\vspace{-0.3cm}

\begin{align}
    \hat{\mathbf{X}} = D(E(\mathbf{X}))
\end{align}

\noindent where $\mathbf{X} \in \mathbb{R}^{N \times M}$ is the input data where $N$ denotes number of data points and $M$ represents number of features, $E$ is the encoding function which maps input data to the hidden layer, $D$ is a decoding function which maps from the hidden layer to the output layer, and $\hat{\mathbf{X}}$ is the reconstructed input (or recovered input). The objective function for autoencoder can be formulated as follows:

\vspace{-0.3cm}

\begin{align}
    \min_{D, E} \left\| \mathbf{X} - D(E(\mathbf{X})) \right\|_2
\end{align}

\noindent where $\left\| .  \right\|_2$ is $\ell_2$-norm. Autoencoder can be used for dimensionality reduction or feature learning when the identity mapping is not desired. That can be achieved either by regularization or hidden layers that learn low-dimensional and non-linear representation of input data. An autoencoder with more than one hidden layer is referred to as a \textit{deep autoencoder}. Each additional hidden layer is a pair of encoder and decoder. A \textit{denoising autoencoder} instead minimizes the following loss function:

\begin{figure*}[t]
        \centering
        \subfloat[RGCN using denoising autoencoder]{
                \centering
                \includegraphics[width=0.7\linewidth,height=4cm]{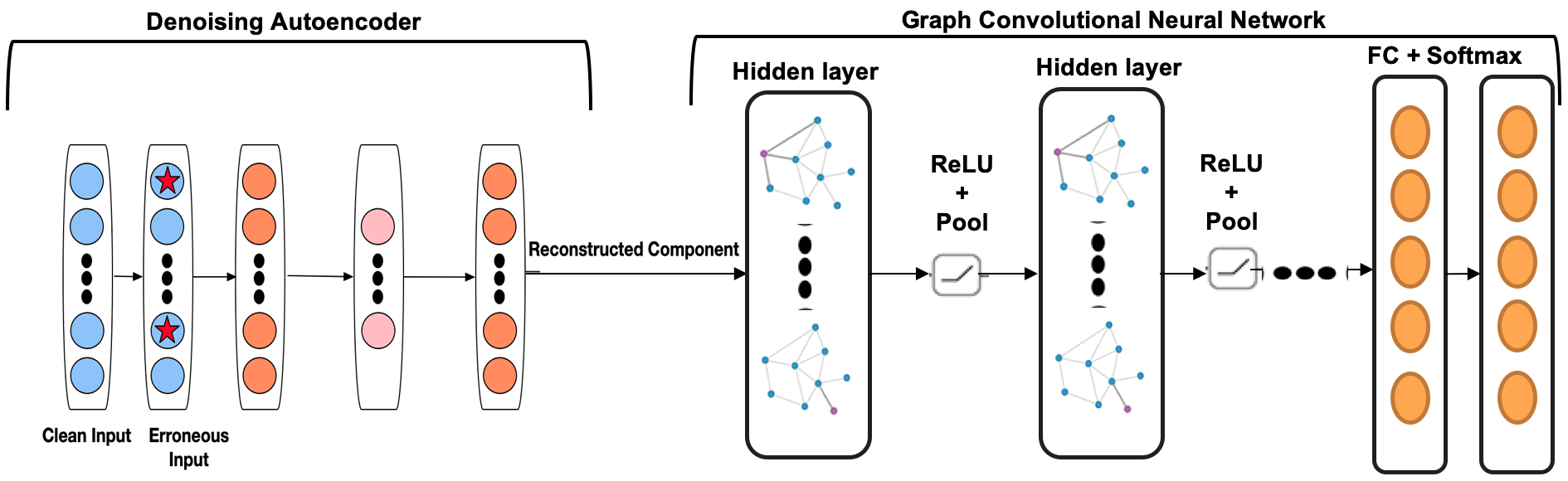}
                \label{fig:first}\hfill}\\
        \subfloat[RGCN using low rank autoencoder]{
                \centering
                \includegraphics[width=0.7\linewidth,height=4cm]{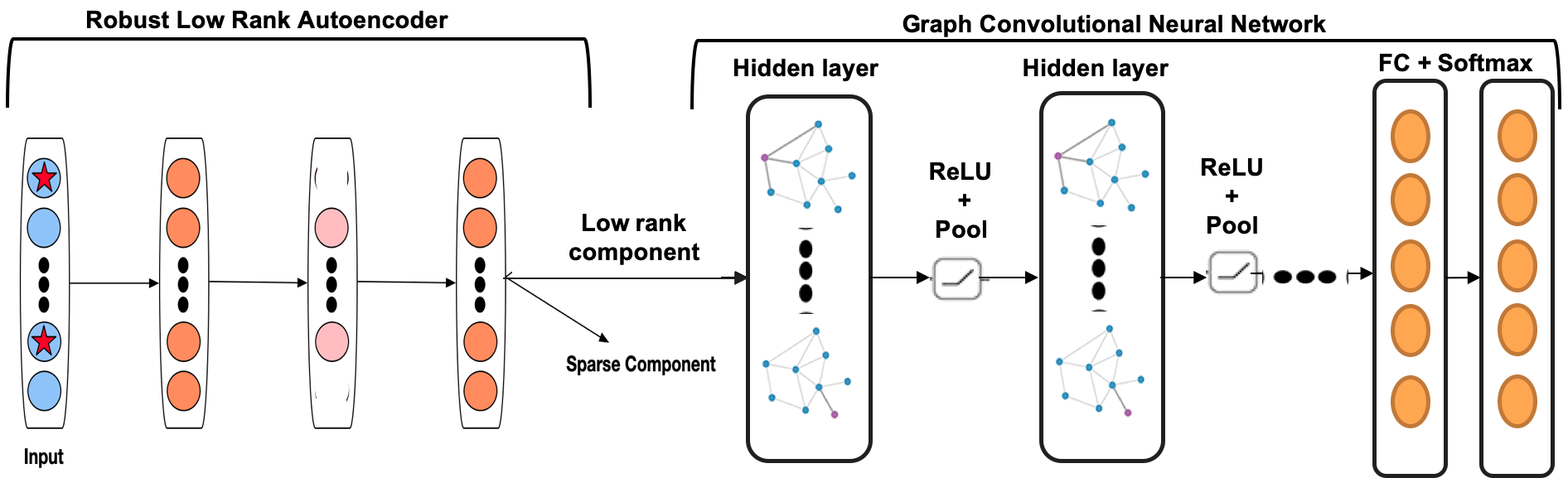}
                \label{fig:second}\hfill}\\
    \caption{Robust graph convolutional neural network for single view data (the schematic representation of one hidden layer for GCN has been extracted from M. Defferrard's website)}
    \label{fig:rgcnsingle}
 \end{figure*}
 
\vspace{-0.3cm}

\begin{align}
     \min_{D, E} \left\| \mathbf{X} - D(E(\tilde{\mathbf{X}})) \right\|_2
\end{align}

\noindent where $\tilde{\mathbf{X}}$ is a copy of $\mathbf{X}$ that has been contaminated by some type of error. The denoising autoencoder is an autoencoder that receives erroneous data as input and is trained to predict the original, uncontaminated data point as its output i.e., reconstructed component. For this reason, the denoising autoencoder has shown to be useful for improving robustness as well as generalizability of the model. 

Motivated by robust principle component analysis (RPCA) \cite{Candes:2011:RPC:1970392.1970395}, an \textit{robust low-rank autoencoder} outputs a low-rank clean approximation of input data as well as sparse error component \cite{Zhou:2017:ADR:3097983.3098052}. The key idea of RPCA is to decompose the possibly erroneous input data into low-rank approximation part and sparse error component. By explicitly accounting for sparse error component to model error in data, the quality of low-rank representation can be significantly improved. Based on this motivation, the loss function for robust low-rank autoencoder is formulated as rank minimization as follows:

\vspace{-0.3cm}

\begin{align}
    \min_{\mathbf{L}, \mathbf{S}} rank(\mathbf{L}) + \lambda \left\|\mathbf{E}\right\|_0
    \hspace{0.2cm} s.t.~ \left\| \mathbf{X} - \mathbf{L} - \mathbf{E}  \right\|_F^2 = 0
    \label{ldae}
\end{align}

\noindent where $\mathbf{L}$ represents low-rank clean approximation or representation of $\mathbf{X}$, $rank(\mathbf{L})$ refers to the rank of $\mathbf{L}$, $\mathbf{E}$ captures error in the possibly erroneous input data $\mathbf{X}$, and $\lambda$ is trade-off parameter. Since $rank(\mathbf{L})$ and $\ell_0$-norm are non-convex, the objective function in Eq. (\ref{ldae}) is an instance of NP-hard problem. One natural way is to replace $rank(\mathbf{L})$ with the trace norm $||\mathbf{L}||_*$ and $\ell_0$-norm with $\ell_1$ norm. The resulted objective function (or loss function) is as follows:

\vspace{-0.3cm}

\begin{align}
    \min_{\mathbf{L}, \mathbf{S}} \left\|\mathbf{L}\right\|_* + \lambda \left\|\mathbf{E}\right\|_1
    \hspace{0.2cm}s.t.~ \left\| \mathbf{X} - \mathbf{L} - \mathbf{E} \right\|_F^2 = 0
    \label{ldae2}
\end{align}

The trace norm is the convex envelope of the rank. As a result, minimization of the trace norm is equivalent to low-rank structure. The $\ell_1$-norm of $||\mathbf{E}||_1 = \sum_{(i, j)} |e_{i, j}|$ is well-known to be a convex surrogate of $||\mathbf{E}||_0$. Algorithm \ref{alg} shows the optimization procedure for the challenging objective function Eq. (\ref{ldae2}) using Alternating Direction Method of Multipliers (ADMM) \cite{Zhou:2017:ADR:3097983.3098052}. The ADMM minimizes one part of the loss function while the others are fixed. More precisely, to obtain $\mathbf{L}$, we fix other variables such as $\mathbf{E}$. We train autoencoder part using backpropagation. To approximate $\mathbf{E}$, we keep other variables fixed. $\mathbf{E}$ can be then solved using proximal operator for $\ell_1$-norm defined as follows \cite{Lin_theaugmented}:

\vspace{-0.3cm}

\begin{align}
    {prox}_{\lambda} (\mathbf{E}) = max(\mathbf{E} - \lambda, 0) + min(\mathbf{E} + \lambda, 0)
\end{align}

\begin{algorithm}[t]
\caption{Robust Low-Rank Autoencoder}
\label{alg}
\hspace{-6.3cm}\textbf{Input}: $\mathbf{X}$\\
\hspace{-5.2cm}\textbf{Parameter}: $\lambda$, $tol$\\
\hspace{-1.7cm}\textbf{Output}: $\mathbf{L}$, $\mathbf{E}$
\begin{algorithmic}[1]
 \STATE $\mathbf{E} = 0$, $\mathbf{L} = \mathbf{X} - \mathbf{E}$ 
 
 \REPEAT
 \STATE minimize $|| \mathbf{L} - D(E(\mathbf{L})) ||_2$ using backpropagation
 \STATE set $\mathbf{L} = D(E(\mathbf{L}))$ which is a reconstruction term
 \STATE $\mathbf{E} = \mathbf{X} - \mathbf{L}$
 \STATE $\mathbf{E} = prox_{\lambda} (\mathbf{E})$
\UNTIL{$\frac{\left\|\mathbf{X} - \mathbf{L} - \mathbf{E}\right\|_2}{\left\|\mathbf{X}\right\|_2} < tol$}
\label{algu}
\end{algorithmic}
\end{algorithm}

\begin{figure*}[t]
        \centering

                \includegraphics[width=0.8\linewidth,height=7cm]{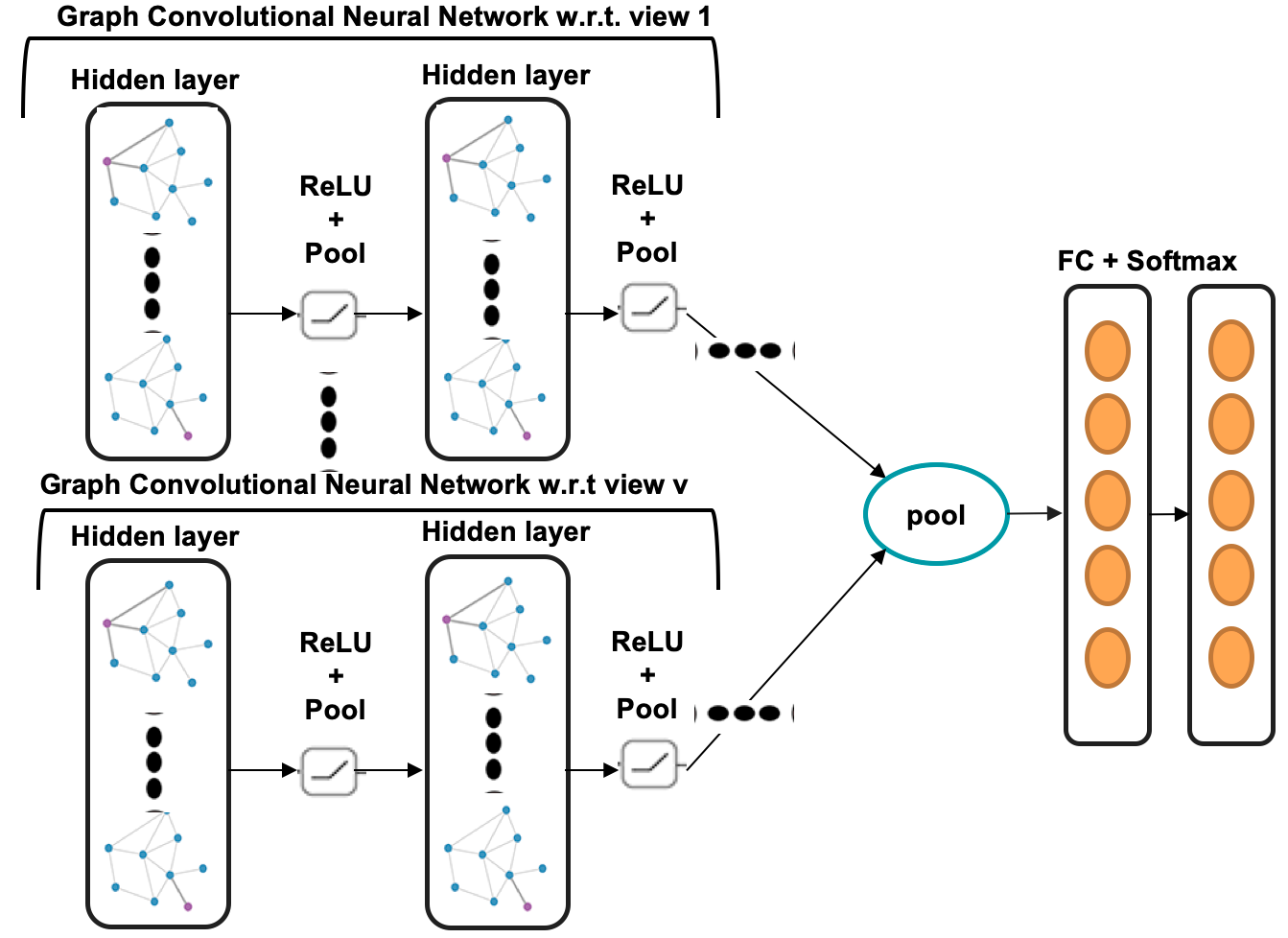}
    \caption{Robust graph convolutional neural network for multi-view data (part of the schematic representation of one hidden layer for GCN has been extracted from M. Defferrard's website)}
    \label{fig:rgcnmulti}
 \end{figure*}
\subsection{Problem Formulation}
 We state the problem for error-robust GCN as follows:
 
\textbf{Problem.} In the setting of supervised or semi-supervised learning, given $N$ distinct possibly erroneous training data points (or samples), the goal is to learn a classifier that predicts label for unlabeled data points effectively and efficiently using GCN such that it is robust against error in data. GCN can effectively capture the nonlinearity of data points and possess strong capability to exploit graph characteristics. The training set is represented as $\mathbf{X} \in \mathbb{R}^{N \times M}$, where $N = U + L$ ($U$ denotes number of unlabeled data points and $L$ indicates number of labeled data points), and $M$ denotes the number of features. For supervised learning, $U = 0$, while for semi-supervised learning, $U > 0$. 

\subsection{Model Architecture}

To alleviate sensitivity of GCN against error in data, we utilize robust autoencoder. With that aim, we introduce three architectures for robust GCN (RGCN) and multi-view RGCN (MVRGCN) as follows:

\textbf{Architecture 1.} In this architecture, GCN receives low-rank clean component of robust low-rank deep autoencoder as input data. The robust low-rank deep autoencoder eliminates error from the possibly erroneous data and returns it as low-rank clean component. Note that the low-rank clean approximation has the same dimensionality as possibly erroneous input data. The loss function for this proposed robust GCN (RGCN) has two main terms: loss for GCN and loss for robust low-rank deep autoencoder. Fig. \ref{fig:rgcnsingle} shows the proposed architecture.

\textbf{Architecture 2.} In this architecture, after cleaning up the input data clean training data, GCN receives reconstructed component of denoising deep autoencoder as input data. Fig. \ref{fig:rgcnsingle} illustrates the proposed architecture. 

\textbf{Architecture 3.} We also propose a novel generalization of RGCN for multi-view data (MVRGCN) where data are represented by multiple views. Each view may be described by an arbitrary type and number of features. With the aim of exploiting complementary and consistent information from multiple views rather than relying on the individual view, we use a separate RCGN with respect to each view and then combine the learned feature matrices using pooling to establish common feature matrix across all views. The pooling layer could be max-pooling, average-pooling, or mixed-pooling that is a linear combination of max-pooling and average-pooling with trade-off parameter $\beta$ i.e., mixed\_pooling = max\_pooling + $\beta$ average\_pooling. The schematic representation of the proposed MVRGCN is shown in Fig. \ref{fig:rgcnmulti}.

\subsection{Model Optimization}
In all of the three proposed architectures, robust deep autoencoders and GCN are trained in a joint manner. For Architecture 1, we optimize the proposed model through Adaptive Moment Estimation (ADAM) \cite{Adam} over shuffled mini-batches. The optimization for Architecture 2 requires joint ADAM and ADMM on shuffled mini-batches. For Architecture 3, we train through ADAM and/or ADMM over shuffled mini-batches depending on the type of robust low-rank deep autoencoders. 
\section{Experimental Evaluation}
To evaluate the effectiveness of the proposed RGCN models, we conduct experiments on real-world datasets and compare RGCN with the following baselines: (1) \textbf{Softmax}. (2) \textbf{Fully Connected (FCk)} denotes a fully connected layer with $k$ hidden units. (3) \textbf{Support Vector Machine (SVM)}. (4) \textbf{Multinomial Naive Bayes (MNB)}. (5) \textbf{Robust Deep Autoenoder (RDAE)} denotes robust deep autoencoders with $\ell_1$ regularization and softmax \cite{Zhou:2017:ADR:3097983.3098052}. (6) \textbf{GCNk} denotes GCN with graph convolutional layer with $k$ feature maps \cite{Defferrard:2016:CNN:3157382.3157527}. 

For multi-view datasets, we compare the proposed MVRGCN model with the following baselines: (1) \textbf{Best Single View (BSV)} feeds the most informative view to Softmax. (2) \textbf{Concatenation (Concat.) - Softmax} concatenates all views and then feeds it into Softmax. (3) \textbf{Concatenation (Concat.) - Multinomial Naive Bayes (MNB)} concatenates all views and then feeds it into Multinomial Naive Bayes. (4) \textbf{Concatenation (Concat.) - GCN} concatenates all views and then feeds it into GCN. (5) \textbf{Concatenation (Concat.) - RGCN} concatenates all views and then gives it as input to RGCN using robust low-rank deep autoencoders.

Similar to \cite{Defferrard:2016:CNN:3157382.3157527}, we use \textbf{Accuracy} as performance evaluation metric. Each experiment is repeated for three times, and the mean of each metric in each dataset is reported. We set learning rate for optimization to $0.001$, learning rate decay $0.95$, and momentum of $0.9$. The batch size is set to be $100$. We apply linear search for graph convolution (or filter) size from $\{10, 12, ..., 32\}$ for Chebyshev, Fourier and Spline, and only report the result for the best option. The pooling layer inside GCN is of size $1$. We only report the results for the proposed RGCN model with denoising deep autoencoders on clean dataset becasue we could not successfully clean up the erroneous datasets.

\begin{table}[ht]
\centering
\caption{Statistics of the real-world datasets}
\label{tab:dataset}
\begin{tabular}{c|c|c|c|c}
\hline
\textbf{Dataset} & \textbf{\# training dp.} & \textbf{\# testing dp.} & \textbf{\# views} & \textbf{\# classes}\\
\hline
20News    & 10171 & 7532 & 1 & 20 \\           

MNIST     & 60000 & 10000 & 1 & 10 \\

Fox      & 1000 & 523 & 2 & 4 \\   

\hline

\end{tabular}
\end{table}

We implement the proposed RGCN and MVRGCN models to investigate their effectiveness: RGCNk(RLDAE) denotes Architecture 1 with graph convolutional layer with $k$ feature maps, RGCNk(DDAE) which represents Architecture 2 with graph convolutional layer with $k$ feature maps. MVGCNk denotes the proposed GCN model for multi-view dataset with graph convolutional layers with $k$ feature maps. MVRGCNk(RLDAE) denotes multi-view GCN with robust low-rank deep autoencoders with graph convolutional layers with $k$ feature maps, while MVRGCNk(DDAE) indicates multi-view GCN with denoising autoencoders with graph convolutional layers with $k$ feature maps.
Statistics of the real-world datasets are summarized in Table \ref{tab:dataset} (training dp. denotes training data points and testing dp. indicates testing data points).

\subsection{Text Categorization on 20News}

Similar to \cite{Defferrard:2016:CNN:3157382.3157527}, the preprocessing steps consist of removing short documents and infrequent words. We then extract 10000 most common words from the unique words across all remaining documents. Each document is modeled as a bag of words and normalized across words. The feature graph is a 16 nearest neighbours, where vertices are those 10000 words, and edges represent the similarity between the words computed based on their corresponding word2vec embeddings using the following formula:

\vspace{-0.3cm} 

\begin{align}
    W_{ij} = exp(- \frac{\left\| vec_i  - vec_j\right\|_2^2}{\sigma^2})
    \label{kernel}
\end{align}

\noindent where $W_{ij}$ denotes edge weight between vertices $i$ and $j$, and $vec_i$ and $vec_j$ indicate word2vec embeddings for vertices $i$ and $j$, respectively. $\sigma$ denotes standard deviation. We obtain pre-trained word2vec word embeddings from GoogleNew-vectors-negative300. Table \ref{table-20news} presents the comparison results on this dataset. The robust autoencoders is of hidden layers with 5000 and 350 units. Although the proposed RGCN models are not superior to MNB on clean 20News (first column), they outperform the fully connected networks which has much more trainable parameters and is thus of higher space complexity.

To evaluate the robustness of the proposed RGCN methods on noise, we add \textit{masking noise} (feature specific corruption) with various levels $\{0.2, 0.4\}$ on 20News training data points (denoted as E20News0.2 and E20News0.4). In masking noise, a set of random features are assigned value $val$. In our experiments, we set $val = 10$. Table \ref{table-20news} also presents the results on erroneous 20News (second and third columns). The proposed RGCN model with robust low-rank deep autoencoders outperforms the baselines which demonstrates its robustness against error of type of masking noise. 


\begin{table}[ht]
\centering
\caption{Comparison results on 20News (mean)}
\label{tab:result1}
\begin{tabular}{c|c|c|c}
\hline
\textbf{Method} & \textbf{20News} & \textbf{E20News0.2} & \textbf{E20News0.4}\\
\hline
Softmax      & 66.28    & 61.45     & 58.12\\
FC2500       & 64.64    & 59.08     & 55.67\\
FC2500-FC500 & 65.76    & 54.87     & 53.12\\
SVM          & 65.90    & 62.07     & 60.00\\
MNB          & \textbf{68.51}    & 64.04 & 62.00\\
RDAE         & 65.70    & 62.90         & 61.05\\
GCN32        & 68.26    & 63.30         & 60.40\\
RGCN32(RLDAE)& 67.90    & \textbf{66.21}        & \textbf{65.31}\\
RGCN32(DDAE) & 67.01    & NA         & NA\\
\hline
\end{tabular}
\label{table-20news}
\end{table}

\subsection{Image Classification on MNIST}

Although the main motivation of the proposed RGCN models is related to graphs with irregular structure, we apply them on image datasets which can be represented as graphs with grid (or regular) structure. In MNIST dataset, each data point in the dataset is a 2D grid of size $28 \times 28$. The feature graph is constructed as an 8 nearest neighbours graph of the 2D grid. The edge weight is computed by the formula in Eq. (\ref{kernel}).

Table \ref{table-mnist} presents the results on this dataset. The proposed RGCN model with robust low-rank deep autoencoders outperforms the baselines on clean MNIST (first column). The robust deep autoencoders is of hidden layers with 392 and 196 units. We investigate the robustness of the proposed RCGN models against error by adding \textit{Gaussian noise} to MNIST training data points. Gaussian noise is one of the popular noise models for image datasets. The mean of Gaussian distribution is set to $0$, while standard deviation (or noise level) is chosen from $\{0.01, 0.02\}$ denoted as EMNIST0.01 and EMNIST0.02, respectively. Table \ref{table-mnist} also presents the results for erroneous MNIST dataset. The proposed RGCN model with robust low-rank deep autoencoders is consistently superior to the baselines and therefore robust against this error type.

\begin{table}[ht]
\centering
\caption{Comparison results on MNIST (mean)}
\label{tab:result1}
\begin{tabular}{c|c|c|c}
\hline
\textbf{Method} & \textbf{MNIST} & \textbf{EMNIST0.02} & \textbf{EMNIST0.04}\\
\hline
Softmax      & 89.78    & 88.09         & 85.14\\
SVM          & 98.02    & 95.00         & 93.01\\
MNB          & 83.31    & 80.00         & 76.50\\
RDAE         & 92.00    & 89.00         & 87.43\\
GCN10        & 97.47    & 95.42         & 93.04\\
RGCN10(RLDAE)& 98.00    & \textbf{97.03}& \textbf{96.00}\\
RGCN10(DDAE) & \textbf{98.03} & NA   & NA\\
\hline
\end{tabular}
\label{table-mnist}
\end{table}

\subsection{Semi-Supervised Classification on Fox}

With the aim of evaluating the effectiveness of the proposed MVRCGN model, we conduct experiments on FOX\footnote{\url{https://sites.google.com/site/qianmingjie/home/datasets/}} which is a small dataset and has two views (view1: text, view2: image). Text view contains text about news, while image view consists of one image about the corresponding news. The preprocessing for text view consists of removing infrequent words, computing TFIDF for 2711 top words. For image view, we normalize the images. 

Since number of features in each view is different, the input data is a similarity graph, where vertices are data points and edge weight is computed by cosine similarity for text view and Euclidean distance for image view. The feature graph for text view is constructed based on Euclidean distance between word2vec embeddings of those 2711 top words. We use Eq. (\ref{kernel}) to compute edge weight for text view. For image view, the feature graph is a grid of the same size as number of pixels in each image. View fusion in the proposed MVGCN and MVRGCN models are done using max-pooling. Table \ref{table-fox} presents the results on Fox dataset. The proposed MVGCN model is superior to the baselines on clean Fox (first column).


\begin{table}[ht]
\centering
\caption{Comparison results on Fox (mean)}
\label{tab:result1}
\begin{tabular}{c|c|c|c}
\hline
\textbf{Method} & \textbf{Fox} & \textbf{EFox0.2} & \textbf{EFox0.4}\\
\hline
BSV                  & 65.19    & 59.87    & 56.02\\
Concat.-Softmax      & 60.83    & 55.64     & 53.00\\
Concat.-MNB.         & 65.40    & 61.56     & 58.20\\
Concat.-GCN32        & 68.60    & 60.00     & 56.45\\
Concat.-RGCN32(RLDAE)& 66.01    & 61.98     & 59.70\\
MVGCN32              & \textbf{72.76}    & 66.00         & 59.00\\
MVRGCN32(RLDAE)      & 70.10    & \textbf{68.02}     & \textbf{64.50}\\
MVRGCN32(DDAE).      & 70.00    & NA     & NA\\
\hline
\end{tabular}
\label{table-fox}
\end{table}

To investigate robustness of the proposed MVRGCN model against error in data, we inject masking noise with various levels $\{0.2, 0.4\}$ to Fox dataset denoted as EFox0.2 and EFox0.4, respectively. Table \ref{table-fox} also presents the results on erroneous Fox. The proposed MVRGCN model with robust low-rank deep autoencoders outperforms the baselines and is thus robust against that error type compared to the other approaches. The proposed MVRGCN model with robust low-rank deep autoendoders is superior to the proposed RCGN model with robust low-rank deep autoencoders whose input data is concatenation of all views. This is mainly because the proposed MVRGCN model exploits complementary and consistent information from multiple views elegantly.

\section{Related Work}
Improving robustness of machine learning algorithms has received considerable attention recently \cite{Najafi2017ErrorrobustMC,DBLP:conf/ijcai/Najafi0Y19,li2016,Pu2019LearningLR}. Existing methods for improving robustness of the deep learning models can be roughly classified into four categories. The first class is to \textit{modify the typical loss function so as to make it error-tolerant} \cite{li2016,Zhou:2017:ADR:3097983.3098052,Pu2019LearningLR,Vincent:2010:SDA:1756006.1953039,Vincent:2010:SDA:1756006.1953039,Xie:2012:IDI:2999134.2999173}.  The proposed RGCN and MVRGCN models belong to this category. There are two broad research directions under this category: (1) training with erroneous input data, (2) training with augmented clean input data. Under the first direction, Zhou and Paffenroth proposed robust low-rank deep autoencoders based on robust principle component analysis for anomaly detection \cite{Zhou:2017:ADR:3097983.3098052}. The idea is to incorporate low-rank decomposition of anomalous data into low-rank clean component and sparse error component in deep autoencoders. The main focus of that work is on anomaly detection. Different from \cite{Zhou:2017:ADR:3097983.3098052}, we feed low-rank clean component of robust deep autoencoders as input data to GCN for robustness improvement. 

As another work under training with erroneous input data, Pu et al. devised robust autoencoders based on robust principle component analysis via disentanglement of two autoencoders to obtain low-rank and sparse components \cite{Pu2019LearningLR}. The evaluation was done for image denoising and music/voice/video separation. Different from \cite{Pu2019LearningLR}, we learn both low-rank clean and sparse error components with one robust deep autoencoders for robust supervised and semi-supervised classification.
The similarity between our proposed models and \cite{Zhou:2017:ADR:3097983.3098052,Pu2019LearningLR} is to incorporate the idea of low-rank decomposition in robust principle component analysis in deep autoencoders to learn clean low-rank component. 

For training with augmented clean input by modifying loss function, the prominent approach is to use denoising autoencoders \cite{Vincent:2010:SDA:1756006.1953039,Xie:2012:IDI:2999134.2999173,Goodfellow:2016:DL:3086952}. Likewise, we used denoising autoencoders in order to robustify GCN, but as we stated earlier, this type of autoencoder requires clean training input data. Therefore, given possibly erroneous input data, filtering is indeed necessary.

In the second category, \textit{extra layer is plugged into the deep learning model} \cite{Han2018,haowang2019,DBLP:conf/iclr/GoldbergerB17}. The proposed RGCN and MVRGCN models belong to this category. One natural existing method here is to add noise layer as first hidden layer to the deep learning model, but it requires clean training data points and labels. On the other hand, there are existing approaches this category that are based on the assumption that input data points are clean, but labels are corrupted \cite{DBLP:conf/iclr/GoldbergerB17,Han2018,haowang2019}. For example, Wang et al. proposed a method for sentence-level sentiment classification given clean sentences but noisy labels \cite{haowang2019}. The key idea is to add noise model layer that learns noise transition matrix from the given noisy labels. Different from \cite{haowang2019}, our proposed models can handle erroneous documents (or sentences) in data, but assumes clean labels. Also, our proposed models can be successfully applied to both text and image datasets. 

In the third category, \textit{erroneous data is filtered out and the deep learning model is trained with the remaining (or filtered out) data points} that should be clean with high confidence \cite{JiangZLLF18}. There are therefore two sequential steps in this category for model training: (1) data filtering (2) model training with clean data points. The proposed RGCN and MVRGCN models with denoising deep autoencoders belong to this category. Jiang et el. presented a novel method for training deep CNNs on corrupted labels \cite{JiangZLLF18}. Different from \cite{JiangZLLF18}, our proposed models can handle erroneous input data, but clean labels. However, please note that we will investigate the robustness of the proposed models on erroneous data with denoising autoencoders in our future work.

The fourth category of approaches uses \textit{adversarial examples} \cite{li2016,Jin2015RobustCN,Zhu:2019:RGC:3292500.3330851}. Adversarial examples leads deep learning models to learn superficial data statistics and causes significant risk for the models deployed in safety-critical systems such as user authentication. They are often created by injecting noise into clean data points. Adversarial training improves robustness of the deep learning models against adversarial examples by training the models with them. Different from this category, the proposed RGCN and MVRGCN models do not explicitly use adversarial training.
\section{Conclusion and Future Work}
In this paper, we developed a novel robust model for single-view and multi-view data via graph convolutional neural networks and robust deep autoencoders. The proposed RGCN and MVRGCN models have several advantages over traditional graph convolutional neural networks and other existing deep learning and non-deep learning models. First, they handle typical types of error with various levels elegantly. Second, an optimization approach via ADAM or combination of ADAM and ADMM shows to converge on several well-known datasets and their erroneous variants well. Compared to the traditional graph convolutional neural networks, some of existing deep learning models such as fully connected networks, and non-deep learning models such as SVM, the proposed RGCN and MVRGCN models achieved better performance on three real-world datasets with typical error models. 

For future work, we plan to apply the proposed RGCN and MVRGCN models on large multi-view datasets such as Reuters. For Fox dataset, we will investigate the effectiveness of the proposed models with word embeddings. Evaluating the robustness of the proposed RGCN and MVRGCN models against outlier-contaminated datasets, and conducting experiments on various architectures for robust deep autoendoers when using two separate deep autoencoders to capture clean low-rank and sparse components are considered as other directions for the future work. It is worth mentioning that the results for Fox dataset does not outperform state-of-the-art methods based on Markov chains e.g., \cite{Najafi2017ErrorrobustMC}. My intuition is that those results are achievable with training the models for more epochs and trying other architectures for robust deep autoencoders.

\bibliographystyle{ACM-Reference-Format}
\bibliography{sample-base}

\end{document}